# Using Deep Learning to Generate Semantically Correct Hindi Captions

Wasim Akram Khan[1]
wasimkhan.dev@gmail.com

Anil Kumar Vuppala
anil.vuppala@iiit.ac.in

Liverpool John Moores University, Liverpool, U.K.[1]

**ABSTRACT**

Automated image captioning using the content from the image is very appealing when done by harnessing the capability of computer vision and natural language processing. Extensive research has been done in the field with a major focus on the English language which gives the scope for further developments in the same with consideration of popular foreign languages.

This research utilizes distinct models for translating the image caption into Hindi, the fourth most popular language across the world. Exploring the multi-modal architectures this research comprises local visual features, global visual features, attention mechanisms, and pre-trained models. Using google cloud translator on the image dataset from Flickr8k, Hindi image descriptions have been generated. Pre-trained CNNs like VGG16, ResNet50, and Inception V3 helped in retrieving image characteristics, while the uni-directional and bi-directional techniques of text encoding are used for the text encoding process. An additional Attention layer helps to generate a weight vector and, by multiplying it, combine image characteristics from each time step into a sentence-level feature vector.

Bilingual evaluation understudy scores are used to compare the research outcome. Many experiments that serve as a baseline are done for the comparative analysis of the research. An image with a score of BLEU-1 is considered sufficient, whereas one with a score of BLEU-4 is considered to have fluid image captioning. For both BLEU scores, the attention-based bidirectional LSTM with VGG16 produced the best results of 0.59 and 0.19 respectively. The experiments conclude that research's ability to produce relevant, semantically accurate image captions in Hindi. The research accomplishes the goals and future research can be guided by this research model.

## INTRODUCTION

### 1.1    Background of the study

Image classification and recognition have been used widely to train models for various applications whereas generating caption turns out to be a challenging task. With the expeditious development of artificial intelligence over the past few years, image captioning has caught the attention of lots of researchers. There are different approaches when it comes to captioning an image, where each caption can focus on different characteristics of an image. Hence it is difficult to come up with a particular caption for the given image. For a model to represent semantic knowledge in natural language, caption creation involves not only the recognition of the items in the image but also relationships between those things, their attributes, and actions.

For those who cannot read the images because of visual impairment, image captions might be helpful. Furthermore, the image descriptions that are generated can be read out using AI to help the visually impaired group thus easing them to interpret the content. It would also be useful for monitoring or scrutinizing the interactions on various social media platforms to keep up with the guidelines. Well-trained captioning models can help the disaster management teams to identify the underlying conditions to respond duly. It can even prove beneficial to the public relations department where a huge amount of data is shared through images. Thousands of images are being generated or uploaded on the web regularly, these captioning models can help generate captions to sort and organize and keep a track of these images.

Image caption generation is now more feasible as the systems can be trained using Neural networks. Earlier experiments involved image captioning using top-down and bottoms up approaches. (Farhadi et al., 2010; Elliott and Keller, 2013). But in recent experiments, the encoder-decoder model performed better than previous methods (Vinyals et al., 2015)(Xu et al., 2015). In this research, the encoders are Convolutional Neural networks and Recurrent Neural networks. Both the CNN and RNN encoders will encrypt text as well as images. The decoder that combines both inputs from the encoders is a neural network model.

With almost 500 million users as its only language of communication, Hindi is the most frequently spoken and understood language in southeast Asian nations. Despite Hindi being the fourth most spoken language, there is a limited amount of research work done on it as data of non-English languages are comparatively less. Methods such as crowdsourcing to assemble data are used by the researchers for their studies. Chinese, Japanese, Spanish and German image captions were gathered, and monolingual image caption models were created. These models or dataset can be used in future research. The caption generation would have a far-reaching audience to cater to. The application of this model would be extensive and would probably have a significant effect on the people utilizing it.

Other most popular languages where extensive research related to image captioning are done are English, Spanish, and Chinese. For English language there are various image captioning dataset, such as, Flickr8K (Hodosh et al., 2013) dataset, has been frequently used as foundational resource in studies of image captioning in various languages. This dataset contains over 8000 images and more than 40000 image description. In order to achieve the objective of the research we would need to generate dataset with Hindi captions. Hence this research will construct the dataset by interpreting the Flickr8K dataset to Hindi language. For the translation, we will be using Google Cloud Platform Translator API.

## 1.2 Problem Statement

There are many potentials uses for image captioning in both English and Non-English languages. Image captioning can be used to classify the pictures people have in their personal electronic devices. Millions of photos can have captions in the user's language on social media websites and applications. In order to inform the proper authorities of scene aspects in a disaster-affected area, satellites may employ well-trained picture caption models. On a regionally localised e-commerce website, it is possible to use the native picture caption model. For those who are blind or are visually impaired, it can also be utilised to provide audio descriptions of the images.

This section sheds light on studies and research done in the field of image captioning. Before this research work, several authors have published and proposed various methodologies and techniques to caption content in different languages. Each subsequent section is further split depending on the methodologies and approaches used in the field of image captioning. The approach used by prominent work has been considered and analysed with reference to the information published by the authors.

There are many potential uses for image captioning in both English and Non-English languages. Image captioning can be used to classify or sort the pictures people have in their personal electronic devices. Millions of photos can have captions in the user's language on social media websites and applications making it easier for the users to understand the pictures while it would be helpful for the displayers like the marketing agencies or other organizations to deliver the correct message to their targeted section of people. In order to inform the proper authorities of scene aspects in a disaster-affected area, satellites may employ well-trained picture caption models. On a regionally localised e-commerce website, it is possible to use the native picture caption model. For those who are blind or are visually impaired, it can also be utilised to provide audio descriptions of the images.

Before this research work, several authors have published and proposed various methodologies and techniques to caption content in different languages. Multiple techniques and various data sets were experimented upon, but very limited amount of work is done in the image captioning for Hindi Language. Hindi as a language is used by almost 500 million people as it is the most common language in the south Asian countries. Models that generate accurate captions in Hindi will help the local people in tackling the issues of interpreting the captions of other languages and then trying to understand the message that is conveyed through the image.

### 1.3 Aim and objectives

Given the paucity of prior research on the subject, this study intends to explain an image in Hindi language by captioning it. Output of encoder-decoder architecture for creating a Hindi caption can be useful in an area of image captioning. In order to develop an encoder-decoder model, this work investigates using pre trained Convolutional neural network to encode picture data and forms of LSTM-RNN to encode text features.

The research objectives are formulated based on the aim of this study which are as follows:

- To provide well framed captions in Hindi for the images that will be very beneficial for a large section of population.
- To analyse the several techniques and deep learning methodologies which can be utilized to achieve image captioning outcomes.
- To compare the outcomes of LSTM and Bidirectional LSTM with different kinds of pre trained CNN and Attention mechanism.
- To evaluate the generated caption this research uses machine evaluation.

### 1.4 Research Question

Most of the image captioning research work has been done in languages like English, Chinese, and Spanish. Hindi is the fourth most spoken language with almost 500 million native speakers. Many people living in the Indian subcontinent use Hindi as the only means of communication. Captioning Images in Hindi will be beneficial for a large group of the population who can understand only Hindi. In a quest to resolve this issue, the doubt arises of How deep learning can be utilized to generate image captions in Hindi which are semantically correct. This research is based on the largely popular dataset, Flickr8k, which is most portable and efficient data set available. The Flickr8k dataset image descriptions are translated to Hindi using GCP translator and used in the research. Our research mainly focuses on generating the best defining captions in Hindi.

### 1.5 Scope of the Study

Numerous research is being conducted on image captioning, but few of them focused on the Hindi Language. The research work provides captions to images in Hindi. The research would facilitate efforts in area of Natural language processing and Computer vision. People researching in other non-English languages can benefit from the methodologies applied in this work. The flickr8k dataset forms the root of the research as it consists of 8000 images with almost 40000 descriptions. This dataset helps to curate dataset with Hindi description which can also be used for further studies for the same language. Although crowdsourcing might have increased the quality of caption generation, it was not used due to time and resource constrains.

## 1.6 Significance of the Study

With the changing technology, visuals have formed an inseparable part of our daily lives. There are millions of images circulated on the web every single day. The significance of image captioning lies in the fact that it makes the media accessible to all. This research could also help to aid efforts of other foreign languages image captioning studies.

The research targets people using Hindi language as their mode of communication. Hindi can also be termed as a language which morphological complex. It can be advantageous for people with visual disabilities. Moreover, the firms trying their hand in the digital world can find image captioning in Hindi as beneficial if their clientele are Hindi speakers. It can be of great advantage with visuals that lack clarity for the users. Image captioning also facilitates in improving Search engine optimisation rankings by making it easier for the page to be crawled by search engine bots.

As well, education is one of the domains where this research can contribute. Hindi native children can be thought various concepts which could prove effective in early childhood learnings. In order to adhere to the rules, it would also be helpful to monitor or carefully examine interactions on various social media platforms. The crisis management teams can determine the underlying conditions to properly respond with the aid of well-trained captioning models.

## 1.7 Structure of the study

The structure of the study are as follows.
- In Section 1, the brief background of the study on Hindi image captioning is presented along with a problem statement. Sections 1.3 address the research Aims and Objectives and present the research's addition to the body of knowledge. Research questions are discussed in section 1.4. The section 1.5 describes the scope and boundary of the research. The significance of the study is provided in section 1.6.
- Section 2 provides insight on Literature review caried out for Image captioning in English and other foreign languages. This section covers the purpose of our study and gaps in previous literature are identified. Also, knowledge and ideas which have already been established on Image captioning in different languages are shared in this section.
- In Section 3, Research methodology is proposed. The section 3.2 contains the details of research design which describes the methodology to be followed throughout the research. In Section 3.3 Tools which are required for implementation of the models are described in detail.
- In Section 4, a detailed discussion about the Implementation of model. The section 4.3 and 4.4 states more about data preparation and processing. In Section 4.5, model development is explained in detail. Whereas, in section 4.6, conversations of different experiments are highlighted.
- In Section 5, the detailed discussion of using different pre-trained CNNs and their results are highlighted. This helps to understand which CNN model will yield the best results. A high-quality caption comparison is also done further with the images to give the study more highlights of the experiments in section 5.2.
- In Section 6 provides details on how this research was carried out and what ways, approaches, and techniques were utilized. Future recommendations to the field of deep learning and image captioning are shared. Contribution of this research and Future recommendations are listed in section 6.3 and 6.4 respectively.

## 2. LITERATURE REVIEW

In our daily life, we encounter tons of images with or without captions. However, being human being, we have the sense of understanding the images even if they aren't captioned. Whereas when it comes to machines, they need certain human assistance to generate captions for such images. Extensive work has been done in the field of automated image captioning since the introduction of Machine learning and Deep learning techniques. Research has been conducted in numerous ways to generate captions and facilitate ease in the fields which can deploy these methodologies. The automated generation of can prove helpful in biomedicine, educations, military, and web searching.

Image captioning done deploying Artificial Intelligence deals with understanding of the image and the bit of information that is provided in the image to generate captions. The machine interacts with the image provided and its interactions to form accurate captions. This is possible with thorough understanding of syntactic and semantic aspects of the language. Earlier traditional deep learning techniques were deployed that exploits hand crafted features to extract the data which is then passed to the classifier namely SVM. The limitation of this method was that it was not feasible for the large and distinct data sets.

Deep machine learning techniques became more popular for the automated image captioning projects as their algorithms can handle vast number of images with diverse features. Since the last decade, ample of work has been done in the field of image captioning deploying the deep learning algorithms as it feasibly handled complexities and challenges of image captioning.

In this literature review, we will explore different techniques which people have applied earlier to generate captions and how helpful they were. In the later part, we will have a brief review of different languages that were explored for caption generation and the paper available on the same. Finally, we will conclude with a brief discussion and future scope in the field of caption generation. Also, some challenges will also be considered which were faced in previous research. Keeping into account of the past challenges we can propose our research methodology this will help to implement the model.

## 2.1    Image Description in English

With recent advancements, research groups that worked on automated caption generation in English have achieved significant results with their models. As the English language data sets were available easily most of the work was done in this language only. Two methods, specifically the retrieval-based method and the method based on language templates, were used for the preliminary work on image captions. These techniques were not enough flexible because they were based on rigid methods (Bai and An, 2018). With the progress of modern Deep Neural Network (DNN) these methods became outdated and were replaced by Deep Neural Network (DNN). Exceptional results were achieved in the field of computer vision when Deep Neural Networks were deployed. Several DNN approaches were adapted as the popularity of image captioning grew. This segment gives a deeper view of the initial work done using DNN approach for captioning the image. The approach and its justification will be provided in this section about the major methodologies applied in image captioning area.

### 2.1.1    Encoder-Decoder Framework

Kiros et al. (2014a) expanded on research conducted, which uses LSTM for encoding phrase, to address all drawback of the neural language model. This method was one the first model which employed the well-known encoder-decoder model in picture captioning study, and it produced noteworthy outcomes. However, Cho et al (2014).'s neural machine translation model served as inspiration for encoder decoder architecture of picture captioning, encoder decoder model is applied to convert content from a language to a different language. In the Kiros et al. (2014b) picture captioning model, Convolutional neural network and Long Short-Term memory are employed, respectively, to encode textual data and image features, a neural language model is utilised to interpret visual elements conditioned on text vectors.

Convolutional neural network was employed by (Vinyals et al., 2015) to encode images and was used as an encoder and RNN-LSTM which translates features of the image into a text using the same idea. According to the encoder and decoder paradigm developed by (Vinyals et al., 2015) picture captioning is defined as estimating the likelihood of a phrase based on an input image characteristic. Donahue et al. (2014) used similar framework of encoder and decoder as (Vinyals et al., 2015)'s study. Text and image features at each time step to the sequential language model rather than just the system's initial stage was supplied by (Donahue et al., 2015)(Mao et al., 2014). Images semantic features were retrieved and these features incorporated to every unit of RNN-LSTM retrieved the throughout the encoder and decoder models image captioning model by(Jia et al., 2015).

### 2.1.2 Multimodal Learning

(Kiros et al., 2014) commenced the work using DNN. Using multimodal log-bilinear this approach first pulls the features from the images and sends it to the neural language model. This model utilizes the multimodal space which aligns the text and image features and determines the conditioned word on the image feature and the previously formed word. Since this model had a drawback of not being able to handle large data, Recurrent Neural Network was applied by (Mao et al., 2014) which is a neural language model. One of the major constraints of the neural language model was cited as its inefficiency to work along a long-term memory.

### 2.1.3 Bi-directional Model

Bidirectional LSTM models can process both past and future information about the text provided to them. Owing to this feature, Bidirectional LSTM models are usually preferred over LSTM models. An image captioning model was developed using pre-trained CNN and Bidirectional LSTM by (Wang et al., 2016). The model achieved a BLEU1 score of 65.5 using flickr8k data, which indicated that the caption generated had a superior quality to the previous models. (Xiao et al., 2019) built the model applying Bi-LSTM and Densely Semantic Embedding Network with a good score of 72.0. Bi-LSTM produced better captions when compared to uni-directional LSTM as it can maintain more information. The research employs both bi-directional and uni-directional LSTM. Some of the features on the images are neglected by the Bi-LSTM model hence, an additional added layer will resolve the limitation.

### 2.1.4 Attention Based Model

The following improved version of framework known as encoder and decoder which is also called an attention guided framework. Xu et al. suggest that of initial attention mechanism in image caption generator. This framework was used in attention mechanism, and the model focuses more on the prominent area of an image while creating an image description. It is a technique that can give various weights to different areas of an image. For instance, it can increase the weights assigned to an image's key area. Attention based model developed by (Xu et al., 2015) involved giving a random area of an image weights. As a result, when creating captions for images, some crucial information was left out. Semantic based attention model that aims on linguistically significant action or objects within image to get over this issue was developed by (Vaswani et al., 2017). The aforementioned attention mechanism forces graphic attention to be engaged for every word, including those that don't explicitly convey visual information. Stop words do not describe the visual entity. A more sophisticated attention mechanism termed an adaptive attention-based mechanism to solve this issue. It automatically chooses whether to depend on the language model or the visual information.

The model selects the area of an image which focuses on if the adaptive attention model decides to pay attention to the visual signal. (Yang, Z. et al., 2016) examined pre-trained CNNs like VGG-16 and ResNet50 using a semantic attention strategy. The research demonstrates that utilising ResNet-50 as CNN yields superior results than VGG-16. Recent years have seen the introduction of shared visual attention mechanism approach of bottom up and top down by (Anderson et al., 2017)

The pooled convolutional feature vector is used to represent the mechanism group of bottoms up of the prominent region of image. On the other hand, top-down technique predicts attention distribution over the visual regions by using task-specific content. With a BLEU-4 score of 0.369, (Anderson et al., 2017)'s cutting-edge technique produced true caption also had greatest captioning in accuracy of images of the MS-COCO. Thus, the method of Encoder-Decoder, Attention mechanism, and Multimodal are three basic ways to caption an image. In the survey of captioning image by Bai and An, assessment of results of each research for the Flickr8K dataset are evaluated to determine best strategy to captioning images (2018). According to this study, captioning an image effectively requires using an attention technique. The attention mechanism basically consists of an encoder decoder structure in which decoder concentrates on certain aspect of the picture or content at every time-interval (Jia et al., 2015).

Even if the attention-based mechanism predominates, adding up attention to the encoder decoder model makes it more complex and requires powerful computing (Bai and An, 2018). With less complexity, encoder decoder architecture produces an impressive outcome (Li et al., 2016). The goal of this study is to provide a Hindi caption for a photograph. Since this topic has not previously been studied, the encoder-decoder framework's output, which produces a Hindi caption, may be useful in the field of image captioning. In order to develop an encoder-decoder model, this study uses LSTM-RNN to encode text features and pretrained CNN to encode picture characteristics. Additionally, attention layer is added which gets the extracted feature from encoder.

## 2.3     Image Description in Non-English Languages

Although many models had been made for Image captioning most of them are in English while research on foreign languages is limited. Some of the captioning models have been generated for Japanese, German, Chinese, French, Dutch, Spanish, Czechia, Hindi, etc. In this section, we will analyse the languages in which image captioning have been done and the techniques implemented by the authors.

### 2.3.1     Image Description in Japanese

Majority of the work is done in English with respect to image captioning. Considering (Miyazaki and Shimizu, 2016) developed a non-English dataset which was first of its kind i.e., known as "YJ caption 26k Dataset" a Japanese dataset. It was like the MS COCO dataset which is based on the captions gathered through crowdsourcing. (Miyazaki and Shimizu, 2016) tried to find a finest way to generate Japanese captions by comparing the three methods i.e., monolingual, alternative and transfer learning. Researchers concluded that the learning method is well suited for captions generation in Japanese.

Likewise, (Yoshikawa et al., 2017) created Japanese caption data "STAIR Caption", which utilized images from dataset known as MS-COCO. "STAIR caption" too collected the dataset from crowdsourcing like "YJ caption 26k" dataset. "STAIR caption" was considered as the major Japanese language image captioning dataset as it contains more than 800,000 captions in Japanese for the entire images (164,062) in MS-COCO. During Comparison of their model training was done with STAIR caption datasets and machine interpreted data, Yoshikawa et al. (2017) found "STAIR caption" model training generated fluent captions than the machine translated dataset. Multilingual Image Captioning model presented by Tsutsui and Crandall (2017) used YJ caption 26K dataset.

(Tsutsui and Crandall, 2017), attempted the multilingual image captioning using the "YJ caption 26K" dataset. A single model was trained for dual languages i.e., Japanese and English by introducing artificial-tokens at start of the sentence to shift language for captioning. They also trained their model for monolingual captioning i.e., in Japanese only. They concluded that the monolingual models reach better accuracy than the dual-language model. Considering the research work done in other languages it can be concluded that an image caption dataset will be required to train model in languages other than English. The prominent ways to collect and develop a dataset (Yoshikawa et al., 2017)) and ((Miyazaki and Shimizu, 2016) is by accumulating from mass sourcing or utilizing machine translated captions.

(Yoshikawa et al., 2017) were competent to reach a specific natural and accurate caption with a crowdsource captions on which the model was trained. Analysing the above work, it was concluded that the performance of the model is more dependent on captions generated through crowd workers. This may lead to different level of accuracy depending on the data deployed. It was also found that model trained in monolingual gave excellent performances than the multilingual image captioning models. Hence this research is focusing on developing monolingual i.e., only Hindi model captions.

### 2.3.2 Image Description in Chinese

(Li et al., 2016) accumulated descriptions of images from Flicker 8k from machine translator, crowd source and human translator. The captions accumulated through crowdsourcing was titled "Flicker 8k-CN". It was observed that the captions accumulated through crowdsourcing from multiple locations across the geographical region resulted in cultural differences. Due to the cultural influences people used different terminology for captioning the images provided to them. So, it was concluded that since the caption is collected across the geographical boundaries it will have the cultural influence of the region in it. Accuracy comparison of the training of the model with machine interpreted caption, human translated caption, also crowd source caption was done by (Li et al., 2016). They discovered that the training of model with crowdsource caption surpassed the human translated caption model and machine translated caption model in terms of accuracy. Reason being the crowd source captions were more natural and had better fluency than the machine generated captions.

Consequently, (Lan et al., 2017) developed Fluency guided architecture by generating easy machine interpreted image captions through altering the non-fluent ones manually. According to their methodology, by using the "Fluency guided framework" model was which trained fared better than the model which was trained using machine translation. To encourage fluency guided research (Li et al., 2016) generated dataset known as COCO-CN, which was the Chinese version of the MSCOCO data which was compiled through process of crowdsourcing. ((Li et al., 2016) as well generated a dataset in Chinese from crowdsourcing which is similar to the work in Japanese ((Miyazaki and Shimizu, 2016),(Yoshikawa et al., 2017)). Much fluent and natural captions were generated by the models trained through crowdsourcing, however it took a huge amount of time and efforts to gather them.

(Lan et al., 2017; Li et al., 2016) also tried to use human beings to generate and edit the descriptions but the model formed was not as accurate as the ones which was trained through machine translation and even demanded much time and money. It was observed by both Japanese (Yoshikawa et al., 2017) and Mandarin (Li et al., 2016) authors that through crowdsourcing they different captioning results due to cultural influences which led to low accuracy and even demanded large amount of time and resources. Given the significant drawback of crowdsourcing the captions and their limited range of accuracy this project deploys machine generated captions to reach better accuracy and generate Hindi Captions for Images.

### 2.3.3 Image Description in Hindi

Some research has been conducted in the area of Hindi image captions; however, it used a range of datasets. Most of these datasets employ machine-translated text to train the model. A deep attention-based architecture was used by (Dhir et al., 2019) to create their Hindi image captioning model. The researchers used the MSCOCO dataset, and Google Translate was used to translate the description. To preserve the descriptions quality, the researchers engaged two human translators to edit it manually and review the captions. Their BLEU1 score came out to be 0.57. Transformer networks were employed by Mishra et al. (2021) to produce Hindi captions. Using Google Translate, the experiment's dataset was produced. The MS COCO dataset was utilised by the authors for translation. The suggested model obtained a BLEU1 score of 62.5, which is rather commendable. The research of captioning image in Hindi and other languages is examined in this section.

Different kinds of deep learning models have been used on a variety of datasets. There hasn't been a lot of study done on Hindi image captioning. Most of the models were built upon the dataset of MSCOCO. Research chose to use the Flickr8k dataset and the encoder and decoder model after analysing these techniques and dataset. For the purpose of extracting image features, the research compares outcomes of unidirectional and bidirectional Long Short-Term Memory with various kinds of pre trained Convolutional Neural Networks.

Assumption from the results of the literature analysis that human assessment is most efficient way to evaluate the automatically produced caption. In this study, machine assessment metric will be the BLEU score. Both human and machine evaluations will be employed as a metric in this investigation. This investigation on Hindi image captioning has applications for computer vision and the field of natural language processing.

### 2.3.4    Image Description for Other languages

Some research work has been done for generating image captions in French, German, Spanish and Dutch languages. (Elliott et al., 2016) developed prototype for multi-language image caption, which simultaneously deployed German and English content against the images provided. IAPR-TC12 dataset a German cation was used to train the multilingual model. "Multi30k" a dataset formed by (Elliott et al., 2016) for the Flicker30k image data for German captioning through professional human interpreters and crowdsourcing. During the research it was found that sentences translated by human translators had the same number of tokens like English dataset while the image descriptions from the crowdsourcing differed in length and the words used.

French descriptions were added by (Elliott et al., 2016) to extend Multi30k dataset. The image captioning models were evaluated using the two methods i.e., machine and human evaluation method. Multilingual picture captioning model and multimodal translation and a were both evaluated by (Elliott et al., 2016) to compare evaluation approach. This research concluded that human evaluation evaluated captions accurately as compared to metrics generated through machine assessment.

(van Miltenburg et al., 2017) presented work to highlight generation of Image captioning in Dutch language. Image description was collected in Dutch language through crowd sourcing, and it was merged with Multi30k Dataset. On comparison of English and French captions with Dutch image description (van Miltenburg et al., 2017)found that due to cultural differences the captions of the images were different. Japanese captions (Yoshikawa et al., 2017) and Mandarin (Li et al., 2016) also supported the similar observation. DIDEC, a Dutch image caption data, released by (van Miltenburg et al., 2017) to provide picture caption in Dutch. (Gomez-Garay et al., 2018) developed a method for the visually impaired people which generated and verbalized image descriptions in Spanish.

It can be interpreted from the assessment above that the researchers were inclined more towards compilation of an image caption. (Elliott et al., 2016), considered assessment of humans manually as the best evaluation method for image captioning. Therefore, our focus is to evaluate manually but because of budget and time constraints, research believes high BLEU scored caption only for evaluation of humans.

### 2.4    Challenges in Image captioning

Despite the amazing developments in neural picture captioning over the past few years, automatic image captioning is still difficult. In this section, we'll talk about how to develop varied, imaginative, and human-like captions by bridging the semantic gap between English and visual images.

The compositional nature of natural language and visual images presents the first difficulty. A captioning system should be able to generalise by assembling things in various settings, even though the dataset comprises conjunctions of a few objects in the context.

Because traditional image caption systems frequently produce descriptions in sequential fashion, where the following produced word changes on both prior generated picture and word characteristic, they frequently lack compositionality and naturalness. This frequently results in language structures that are syntactically accurate but semantically meaningless, as well as a dearth of diversity in the output captions. With a context-based Attention model, that enables the system to construct sentence which are based on segments of the examined visual picture, it is proposed to overcome the compositionality issue. We specifically used a recurring language model with gated recurrent model, which allows an option to attend to any of textual or visual inputs from previous generation steps at each producing step.

The use of generative adversarial networks (GANs) (Goodfellow et al., 2014) in training the captioner is another innovation we make to focus on the problem of shortage of naturalness. Using GANs, an attention differentiator counts the "naturalness" of sentences and loyalty to the picture based on attention model which matches portions of the images and the text produced and vice versa. The probability of created captions based on attributes of the image and vice versa are scored by the attention differentiator to determine the caption quality. Keep in mind that this counting is at the image pixel and word level, not the level of the entire depiction. In order to accurately capture the compositional aspect of language and visual images, this location in the score is crucial.

The dataset bias that has an impact on the present captioning systems is the second problem. The problem with these systems is that they have difficulty generalising to situations when the same objects appear in scenes with unknown contexts because the models which are trained, overfit to the familiar image object which occur in a context (e.g., forest and bed). Although eliminating dataset preference is a difficult, open research subject in and of itself, analytical tool to measure how subjective a particular image captioning model can be.

The examination of the calibre of created captions presents the third difficulty. Utilizing automated measures, while somewhat beneficial, is still inadequate because they ignore the image. When scoring various and descriptive captions, their scoring frequently remains insufficient and perhaps even misleading. Human assessment continues to be the gold standard for rating captioning systems. In our Turing test, human judges were asked whether a given caption was created by a machine or was real. Most of model generated descriptions were considered as an authentic by human judges, showing that the proposed captioner performs well and appears that potential innovative method for captioning an image.Computer vision models will become further dependable for utilization of a individual helpers for people who are visually impaired and enhancing people's daily lives as progress is made in automatic picture captioning and scene understanding. The difficulty of connecting language and visual on a semantic level highlights the need to enhance scene comprehension with logic and common sense.

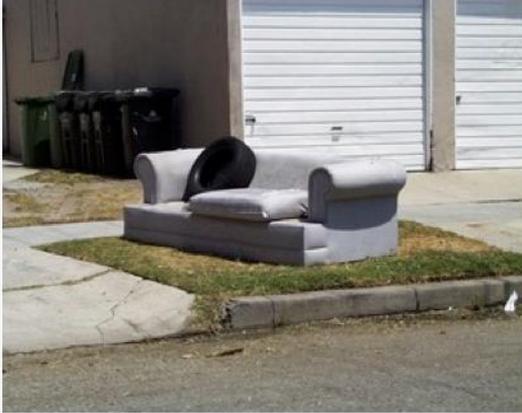
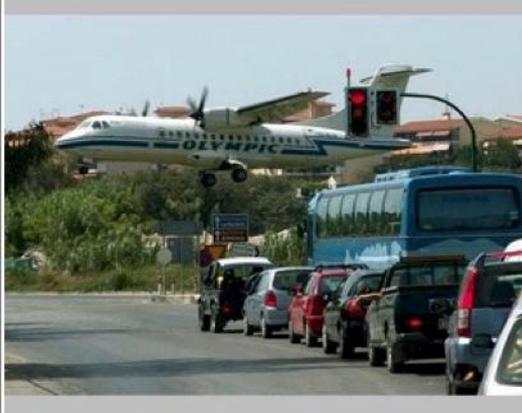

GAN: a couch sitting in front of a house with a trash can
CE: a white couch sitting in front of a house
RL: a couch sitting in front of a house
GT: a white couch on top of a grass curb with a black table in the background

GAN: a large passenger jet taking off from a busy street
CE: a large passenger jet sitting on top of a runway
RL: a group of cars parked on the runway at an airplane
GT: an airplane descends very close to traffic stuck at a red light

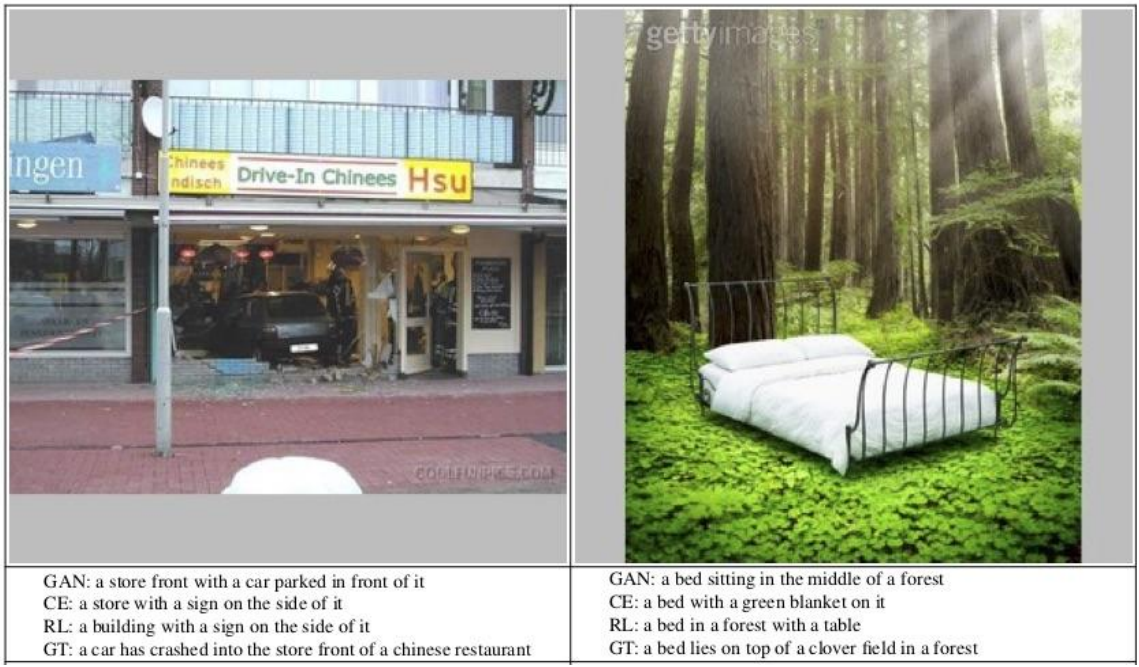

*Figure 1: Out of context images.*

## 2.5 Discussion

Using computers to construct grammatically sound sentences that accurately describe a picture is known as image captioning. While describing an image is a straightforward task for humans, doing so with computers is challenging. Image caption creation is now a simple process thanks to advancements in the field of neural networks. Many deep learning models have been employed to generate captions from photos. Using an encoder-decoder strategy is one effective way to create these captions. The hardest part of this process is figuring out how well we can characterise these photographs. In comparison to past methods, encoder-decoder paradigm was implemented by (Vinyals et al., 2015) by much more effectively. In this system, the encoders consist of Convolutional Neural Networks and Recurrent Neural Networks. Encoding images using the CNN encoder, and text is encoded using the RNN encoder. The decoder that combines both inputs from the encoders is a neural network model.

In comparison to other foreign languages, English has a substantially higher availability of datasets for building these models. Useful English-language statistics include Flickr8k, Flickr30k, and MSCOCO. There hasn't been as much research in the domain of foreign languages because there isn't much information available. Crowdsourcing has been utilised by researchers to acquire data for their studies in several foreign languages, including Chinese ((Lin et al., 2014)), Japanese ((Yoshikawa et al., 2017)), and German ((Elliott et al., 2016)). Researchers created the monolingual picture captioning model to be used in next studies.

There are many potential uses for image captioning in both English and native tongue. You can use image captioning to organise and classify the pictures you have on your PCs and mobile devices. On social media platforms, which are getting more and more popular, millions of photographs can be described with captions in the language of the user. In order to inform the proper authorities of scene aspects in a disaster-affected area, satellites may employ well-trained picture caption models. On a regionally localised e-commerce website, it is possible to use the native picture caption model. For those who are blind, it can also be utilised to provide audio descriptions of the images.

Two methods have been used in earlier works to tackle the captioning problem: a top-down method (Bahdanau et al., 2014) (Sutskever Google et al., 2014) and a bottom-up method (Farhadi et al., 2010) (Elliott and Keller, 2013). In the bottom-up method, words are generated from the input image, and additional texts were added to create captions. For top down method, the image used as an input is turned into a text depending on specific principles. The top-down method uses an end-to-end formulation from an image to a sentence, and all the network parameters are discovered during training. Nowadays, endwise learning is achieved through encoder and decoder systems using the top-down technique. The use of an encoder-decoder system can be used to overcome image captioning issues, as demonstrated by recent developments in neural machine translation employing deep learning (Vinyals et al., 2015). In (Sutskever Google et al., 2014), authors created sequence deep learning machine interpretation models using two LSTMs: one as a decoder to produce the target sentence, and the other is an encoder which encodes the input words. A machine translation model based on encoder-decoders was created in (Bahdanau et al., 2014), which enhanced working which is done by enabling model to inevitably choose parts of the word that are crucial for predicting the target word. By maximising the conditional likelihood, focused phrase is produced to the input phrase in (Cho et al., 2014), where researchers worked two RNN as encoders and decoders.

Multimodal recurrent neural network used for producing the description is the model for caption creation that the authors suggested in (Mao et al., 2014). Using prior words and an image, they created a term using a probability distribution model. Here, the probability distribution is used to construct the caption for the input image. In (Karpathy and Fei-Fei, 2014), the alignment model has been utilised to combine CNN and RNN across picture areas. Here, phrases have been processed by a bidirectional recurrent neural network, and a multimodal embedding is employed to form an objective function that aligns two modalities. The initial encoder and decoder-based model, using encoder is a CNN and decoder is RNN, has been suggested (Xu et al., 2015).

In order to gather the characteristic highlights and feed them into the RNN so they can understand the high-level attributes, authors in (Yang, Z. et al., 2016) extended encoder and decoder model and developed a review model network. Also, authors of (Xu et al., 2015) have created an attention-based model which can dynamically serve various areas of the picture whilst creating image captions. In sequence model which are built recently research has focused primarily on RNN and CNN, which are employed as the encoder-decoder. In order to obtain superior performance, attention-based technique that is well-versed in encoder and decoder design. Based on based on attention mechanism a transformer design without RNN and CNN was proposed by authors in (Vaswani et al., 2017). It has been demonstrated that this model, which has more parallelism and greater accuracy than another machine translation model, is of higher quality.

Transformer networks and Models which are pre-trained language, respectively, were employed the encoder-decoder in (Zhou et al., 2020). (Cornia et al., 2019)describes a meshed-memory transformer architecture that learns multiple levels of depiction of the link amongst picture and ingested previous information to enhance image encoding and language synthesis. A double generative network ensembles generation and approach based on retrieval for picture captions has been developed by authors in (Liu et al., 2020).

Along with English, Hindi is one of India's two official languages. It is extensively used throughout South Asian countries. Hindi is a language that is spoken by more than a billion people worldwide. Therefore, creating a Hindi image captioning model is crucial for the welfare of the populace. To the best of our knowledge, the Hindi language has only one existing work on image captioning (Dhir et al., 2019). However, this suggested solution has the issue of inaccurate caption generation because the model based on language has long-term dependencies. GRU is employed in this for language modelling; it can only partially address the issue of long term reliance. In this study, we address the issue of Hindi picture captioning using a transformer-based paradigm (Vaswani et al., 2017).

## 2.6 Summary

In this section, the methods and approaches used in earlier image captioning studies are in-depth explored. There are several methods for describing an image's content in English or other languages, but no research has been done on how to do it in Hindi. As a result, Hindi image captioning is thought to have a significant research deficit. The team decides to use an encoder decoder model used for Hindi caption an image after examining previous methods. The encoder decoder deep learning model needs the Hindi dataset to be trained, but there is no Hindi-language picture captioning dataset. Due to time and financial constraints, this project translated the English caption dataset into Hindi using machine translation in order to collect the Hindi description. The model which evaluates with machine interpreted sentences has also produced impressive results in earlier work. One of the key conclusions of the picture captioning research is that the optimum method of evaluation is manual observation. As a result, this study also uses human assessment. BLEU utilised as an evaluation which is a machine-based metric in addition to the human evaluation approach.

To sum up, the monolingual learning model was used in each studies of Non English picture description. (Yoshikawa et al., 2017) and (Li et al., 2016) noted differing outcomes in multi-language models for Japanese and Chinese captioning. Their study suggests that there is still room for improvement in captioning of multi-language image in other languages. According to this earlier research, there is room to do several trials to produce image descriptions in the Hindi language. The outcomes of this investigation may help.

## 3. RESEARCH METHODOLOGY

In this section all the details related to research implementation are addressed. Research methodology provides research validity and offers reliable scientific conclusions. Additionally, it offers a thorough plan that aids in keeping researchers on course, facilitating a simple, efficient, and manageable approach. This will help reader to comprehend the strategy and procedures utilised to arrive at results by understanding the researcher's methodology.

This section will provide more aspects of addressing objectives of the research by specifying the methodology that will be utilized. Also, the data understanding, and data preparation strategy are discussed in detail.

This section will discuss the research approach that was employed. CRISP-DM (Cross-Industry Standard Process for Data Mining) is a method that can be used. This method is distinguished by the deployment layer. The CRISP-DM deployment layer is offered. In this study, the CRISP-DM methodology will be used. Domain knowledge, data understanding, data preparation, modelling, evaluation, and deployment are the five processes that make up the CRISP-DM strategy.

Thus, it is necessary to process and change both the visual layer and the linguistic layer of data. Both the image and the text feature are converted into mathematical vector before being fed into the model of image captioning. The model is then assessed and understood by utilising evaluation metrics or by looking at the results.

The blend of CNN and RNN could result in a strong and captioning of images, as previous research has successfully demonstrated. As a result, the encoder-decoder model in the proposed study uses the CNN-RNN paradigm. In their experiment, (Vinyals et al., 2015) discover that pretrained CNN enhances the model's performance. This research utilizes three different pre-trained Convolutional neural networks of Inception-V3, ResNet-50, and VGG-16 to choose the optimal CNN for captioning in Hindi language, drawing inspiration from that study. LSTM is utilised as a recurrent unit in RNN and has demonstrated remarkable success in the field of picture captioning (Bai and An, 2018).

## 3.2 Methodology

The Research methodology to be used in this research is CRISP-DM (Cross-Industry Standard Process for Data Mining). An open standard process model called the CRISP-DM describes typical methods employed by data mining professionals and this is one of the most popular analytics models. Data mining is broken down into six main processes by the CRISP-DM, including business understanding, data understanding, data preparation, modelling, evaluation, and deployment. The phases' order is not rigid and switching back and forth between them is frequently necessary. The research will follow an iterative process and cycle to achieve the goal of study.

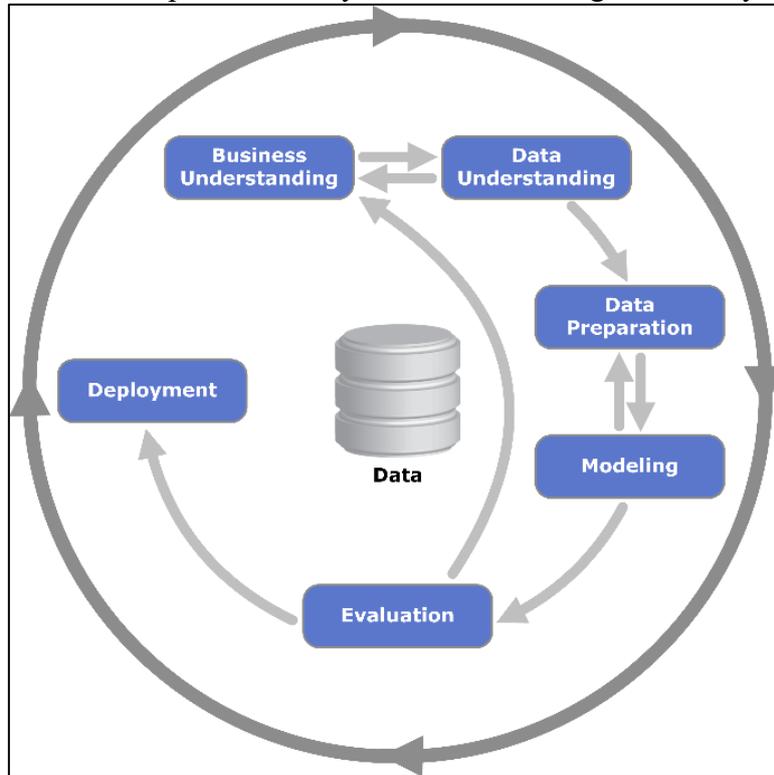

*Figure 2: Cross-Industry Standard Process for Data Mining*

### 3.2.1 Domain Understanding

The literature on Hindi picture captioning is examined in this section. Different kinds of deep learning models have been used on a variety of datasets. There hasn't been a lot of study done on image captioning in Hindi language. Most of the models were built upon the MSCOCO dataset. The project will employ the Flickr8k machine-translated dataset and the encoder-decoder framework after analysing the techniques and dataset. For the purpose of extracting image features, the research will compare the outcomes of LSTM and BiLSTM with various kinds of pre-trained CNN. We can confidently deduce from the results of the literature research that BLEU (Bilingual Evaluation Understudy) is the best technique for assessing the automatically generated captions.

The primary goal of the study is to provide a Hindi visual description. To accomplish the objective, it is essential to collect data on Hindi image captions. In order to create a Hindi version of the Flickr8K dataset, this project translated the English description of the dataset into Hindi using the GCP Translator API.

### 3.2.2 Data Understanding

Understanding the data is crucial to the entire model-building process. We can create models that are more accurate by comprehending the facts. This investigation uses the dataset named as Flickr8k (Hodosh et al., 2013). This dataset is comparatively not that big. Thus, the model can be easily trained on desktops or laptops that aren't very powerful. Additionally, data is correctly labelled, and five descriptions are given for each image. The crucial quality of the Flickr8k dataset is that it is cost-free.

There are 8,000 photos with five different captions, each describing the key elements and actions in the image, make up this new benchmark collection for sentence-based image description and search. Flickr8k text comprises text files describing train dataset, test dataset. Flickr8k.token.txt contains five captions for every image which is more than 40000 captions. The pictures were hand-picked to represent a diversity of events and circumstances from six different Flickr groups, and they usually don't feature any famous individuals or places. We used machine-translated text in this study because Hindi captioning will be the target language. This dataset's English description will be interpreted by GCP Translator API.

For the research, four different kinds of datasets will be generated for training the models. With two datasets including five image descriptions and two datasets comprising one description per image, different variants of clean and unclean datasets will be constructed. The dataset's impure five sentences version uses sentences that will simply translated without any kind of cleaning. In this study, only data from five impure sentences will be used.

### 3.2.3 Data Preparation

This segment describes prior to the data that would be sent into the model to train, it would be pre-processed and converted into relevant training data. For the research, we will be using two sets of data i.e., Image data and text data.

Image data must first be transformed into a vector format for processing before being fed into a deep learning model. In their assessment of deep neural network image captioning models, (Hossain et al., 2018) discovered that most recent state-of-the-art research depend on pre-trained Convolutional neural network model. These models have vectors of fixed size. There are several different pre-trained CNN models that can be used to generate image captions. AlexNet, DenseNet, ResNet50, VGG16, InceptionV3, and SquezeeNet and are some of the highly admired pre-trained Convolutional neural networks. These models are utilized in transfer learning research. The last layer of the Convolutional neural networks is omitted when utilising pre-trained CNN as it is widely utilised in picture classification studies. The penultimate layer, which yields the characteristics of image, will be employed in this study. In order to save time during model execution, obtained features will be stored from these Convolutional neural networks in a pickle file. This could be used again rather than going again through the entire extraction procedure.

The vocabulary size and description length for the impure five sentences in the Flickr8k dataset will have 8,652 and 39 respectively. The dataset would not include any stop words. The image description may become less fluid if stop words are eliminated (Lan et al., 2017). The text data processing stage removes numbers and punctuation from the text.

To instruct the machine where a sentence begins and ends, the text data would be enveloped with start and end markers. Neural network models cannot process text directly. To resolve this issue, the textual data are transformed into integral tokens. In this study, the pre-processed picture descriptions are transformed using Keras tokenizer into integer tokens and are transformed into floating-point values. Zero padding is added to maintain the size of these vectors to be identical as of the image vector. This makes it simple to combine the tokens and treat them as input for the model's decoder.

### 3.2.4 Modelling

In this section will have a look for the modelling approach in this research. CNNs are easier to use, offer better results, and take less time to train, the study uses them. Pre trained Convolutional neural network yields more accurate results as compared to CNN that are constructed from scratch. The scene of the image cannot be used in a created text after the image's features have been extracted. This issue is resolved by the unidirectional LSTM. The precision of the generated text is restricted to short sentences. Bidirectional Long Short-Term Memory Networks are utilised to overcome this restriction. Longer sentences work well with BiLSTM since it stores more data.

It is crucial to choose an appropriate construct of encoder-decoder architecture. (Tanti et al., 2017) concluded that the Merge model is most suitable model for encoder-decoder architecture after comparing 16 alternative models in their research. The merge model can be shown in Figure 3. The word sequence and image feature vectors are both inputs to the merged model. The decoder model then creates the following likely word in the sequence using the blend of two input vectors. Particularly while dealing with image and text, model performs optimally well.

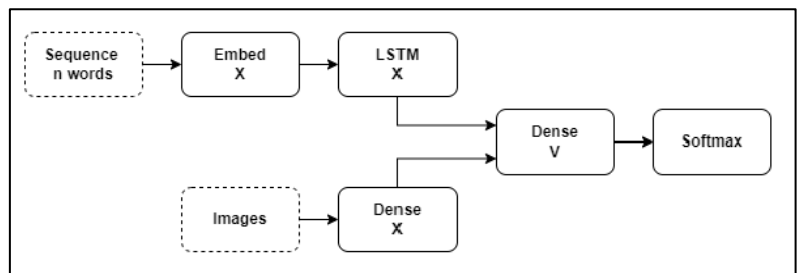

*Figure 3: Merge Model*

RNN-LSTM is the ideal neural network model for handling text data (Tanti et al., 2017). In this study, phrases are encoded using unidirectional and bidirectional LSTM, while visual data is encoded using pre-trained CNNs. There are several techniques where such inputs can be encoded. To add, concatenate, and multiply are the few techniques. These encoded outputs from picture and text data are combined. The optimal method for combining the encoded input is addition. As a result, we added to vectorize the image and text. The decoder, which uses a combination of text and image vectors, is located in the dense layer. SoftMax, the final layer, performs greedy search to yield single word based on the likelihood of the word that came before it and the generated picture feature.

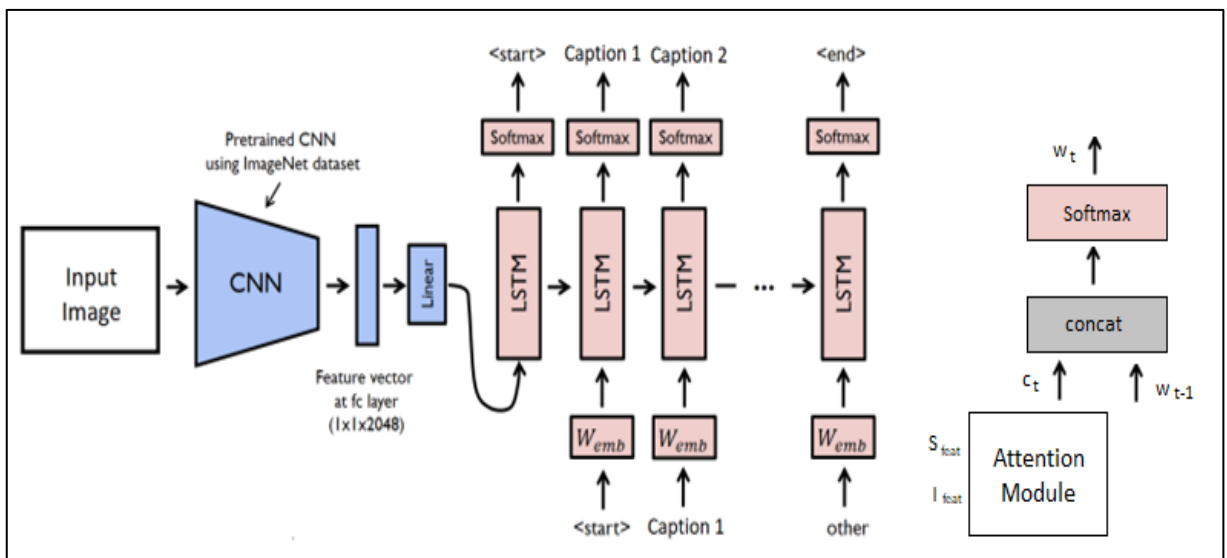

*Figure 4: Image Process Flow*

Figure 4 depicts an encoder-decoder model's process flow. The model is a synthesis of models from CNN and RNN. The model is trained using two inputs: first, an image, and second, associated captions. Each LSTM layer receives a word to forecast, and this process is how the LSTM model learns from captions and improves itself. To extract picture features, pre-trained CNN is employed. Direct correlation between captions and image features is possible using these features. The model creates dataset captions for the final layer that are the longest possible. The vocabulary's length determines the size of the final layer.

Recently, Attentive neural networks have shown promise in a variety of tasks, including image captioning, speech recognition, machine translation, and question answering (Bahdanau et al., 2014). The attention mechanism has numerous significant purposes in various NLP domains, including machine translation and speech recognition. Accuracy in numerous application areas can be further increased by combining Bi-direction LSTM and the attention mechanism. Combining attention mechanism with Bi-LSTM can further enhance the performance of image captioning as attention mechanism has the specific benefit of learning significant information from contextual information.

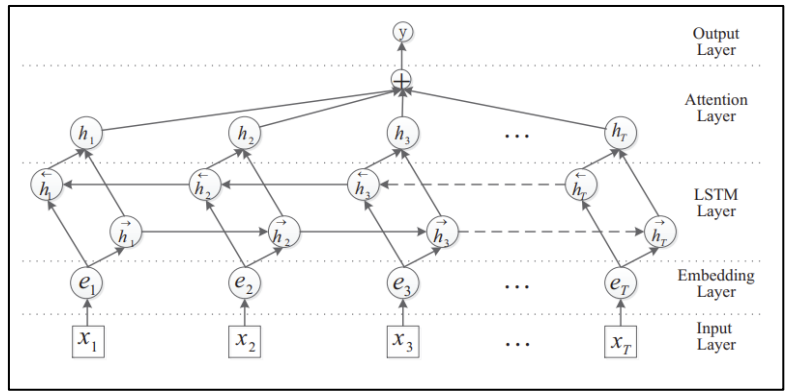

Figure 5: Bidirectional LSTM model with Attention Layer

As shown in the Figure 5, an attention layer is added that would generate a weight vector and, by multiplying it, combine image characteristics from each time step into a sentence-level feature vector. The research will compare results from LSTM, Bi-LSTM, and Att-BiLSTM approaches for image captioning in Hindi.

### 3.2.5 Evaluation

In this study, the created caption is evaluated automatically. The measure BLEU (BiLingual Evaluation Understudy) is used to evaluate machine interpreted text automatically. The resemblance of the machine-translated text to a collection of excellent reference translations is gauged by the BLEU score, which ranges from zero to one. Several 0 indicates that there is little to no overlap between the machine-translated output and the reference translations (poor quality), while a value of 1 indicates perfect overlap between the two translations (high quality).

It has been demonstrated that BLEU scores and human evaluations of translation quality are highly correlated. Keep in mind that even human being interpreters cannot produce a score of 1.0. Sentences that have been automatically translated are evaluated using BLEU (Bilingual Evaluation Understudy). A BLEU score can be used to determine how closely related the reference and candidate sentences are, as well as forecasted sentences. It evaluates the generated description's quality in relation to several reference explanations by calculating the number of n-gram occurrences. The resulting description is more fluid the higher the n-gram score (Papineni et al., 2002). The most typical n-gram size is up to four. In this study, one to four n-grams are employed. BLEU-4 assesses the generated description's fluency, whereas BLEU-1 assesses its sufficiency. Additionally, it assesses the produced explanation's precision, which is measured as the proportion of the candidate sentence's total word count to the number of words that overlap. The fact that BLEU is language-independent was one of the primary justifications for employing it in this study.

It is strongly advised against attempting to compare BLEU results across several corpora and languages. It can be very deceptive to compare BLEU ratings for the same corpus but with different amounts of reference translations. However, the following interpretation of BLEU scores (given as percentages rather than decimals) may be useful as a general guideline.When used to assess single sentences, the BLEU metric does poorly. For instance, despite capturing much of the meaning in both examples, both phrases receive a very low BLEU ratings. BLEU is by design a corpus-based metric, meaning that statistics are aggregated over an entire corpus while computing the score. This is because n-gram statistics for individual phrases have less significance. The BLEU metric described above cannot be factorised for single sentences, as should be noted.

A dropped function word like "a" receives the same penalty as if the name "NASA" were mistakenly changed to "ESA" since the BLEU measure does not differentiate between content and function words.Polarity of sentence can change with the deletion of a single word, such as "not." Additionally, just considering n-grams with n4 disregards long-range relationships; as a result, BLEU frequently only assesses a minimal penalty for grammatical errors. Both the reference and candidate translations are normalised and tokenized before calculating the BLEU score. The final BLEU score is substantially influenced by the normalisation and tokenization methods chosen.

### 3.3    Logical flow of the system

Figure 6 below depicts the design process of the project for captioning of an image. It provides visual evidence of how this project's caption creation procedure operated. Since the descriptions used for training are included in the middle of the tags startseq and endseq, the model can predict captions that begins with startseq and ends along with endseq. According to (Vinyals et al., 2015), the model suggests that a word will be prepared based on the order of the words that have come before and the image vector. In other words, using the provided caption prefix and picture feature, then this model creates a different caption phrase by phrase. The model receives together the image and caption vector, and which forecasts words with the maximum likelihood. This anticipated word gets added to image caption. Hence, model is then given the same word once more.

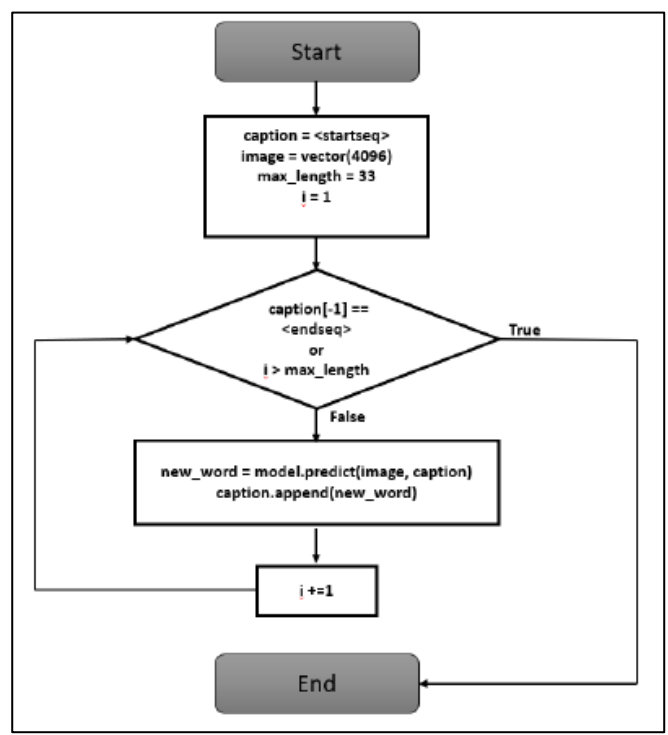

*Figure 6: Design Process flow of the project*

As shown in the Figure 6, The process is iterative and will be split into two parts. An artificial tag with the prefix "endseq" is the first technique to terminate the caption production. The model stops forecasting the following word when this predicts "endseq" tag because it is the final word that emerged in every training description. Giving a maximum length for the created description as a restriction on the caption's length is the second technique to halt caption generation. The model ends and outputs the generated caption in one of the two methods.

### 3.4 Tools

This segment considers about resources needed for research implementation. Moreover, actions engaged for setting up the environment for training the model are also discussed.

### 3.4.1 Python

For model development Python 3.8 version will be employed. Python, a high-level language, is designed for highlighting the readability of codes using important indentations. It is also garbage collected and a language with dynamic typing, where the interpreter decides what type to give variables based on their current values at runtime

### 3.4.2 Keras

The model will use the Keras Library. Keras, an open-source library, which offers a Python interface for artificial neural networks and provides TensorFlow library interface. Few backends such as Theano, Microsoft cognitive service, and TensorFlow is aided by Keras.

### 3.4.3 Google Colab

For detailed project implementation Google Colab is used. Colab notebooks lets blend feasible code and text in particular document, which also could consist of pictures, Hypertext markup language, LaTeX and others. For implementation, Colab will be leveraged for using GPU - Tesla P100. Random access memory can vary from 8 GB to 16 GB

### 3.5 Summary

In this section we discussed how we addressed different research objectives to implement the Image caption in Hindi. This is an incremental approach so the content of research methodology will be updated throughout the research. Data understanding and preparation were reviewed, and process workflow were brought forward. The Modelling techniques were also highlighted in this Section. How the evaluation of the model will be done are presented in detail in this section.

Detailed description of CRISP-DM methodology which was discussed in this section. Past methodologies were also discussed and any challenges in them were highlighted to overcome.

### 4. IMPLEMENTATION

This section provides a thorough explanation of the implementation of Hindi image captioning model and implementation method. This covers all the specific information and supporting data from the beginning to the end of implementation. This section closely focuses on the technical steps used for the approaches and experiments. In foresaid section, different stages of implementation consist of environmental settings, data preparation, pre-processing of data, model training and experiments.

For the development of image captioning in Hindi data plays an integral role. The methodologies for data extraction, data creation, and translation are described briefly in this section. Data used for this research are classified into two, Image data and text data, as Flickr 8k dataset consist of these two datasets. These datasets are pre processed and further transformed for the training of the model. Also, text data should be wrapped with tags and tokenized after pre-processing. A rigorous set of experiments are carried to identify the optimal hyperparameter tuning. Thus, the model is trained with optimal parameters to assess the model's effectiveness.

Details related to setting up the environment are discussed further in this section. These details also include the libraries and pre-trained convolutional neural networks used for the implementation. Keras library was used to run the models. CNN models such as Inception-V3, ResNet-50, and VGG-16 were imported using the Keras library.

This section closely focuses on the approaches and experiments carried out for the research. There are different approaches which were applied in this research. Among these, few approaches were finalized by having multiple experiments to gain satisfying and better results. Several experiments with CNN and CNN-BiLSTM models were carried out to derive BLEU score to evaluate the best model. In this study various experiments with different models are employed but the structure of the best model is highlighted which produces the better quality of image captions in Hindi.

## 4.2    Environmental settings

Implementation of the project is carried out by Google Colab. Tesla P100, a potent GPU with up to 16GB of RAM, is provided by Colab. The processing of data and model building are done using Python 3.7. The Modelling of the picture caption model mainly relies on memory (Vinyals et al., 2015). As a result, all of the trials are run using 128 GB of RAM. This study also utilized elevated level APIs like Google Cloud Translation and Keras for model building and interpretations. Three key advantages of using Keras library are: 1) Quick experimentation and simple comprehension; 2) Support for CNN, RNN, and their hybrids; and 3) CPU-based operation.

## 4.3    Data Preparations

The given dataset is in English; however, the proposed effort intends to label an image in Hindi. Two techniques could be used to create Hindi captions: first is to gather Hindi picture descriptions from the of Hindi speaking community, and the other is to use machine translation to convert the dataset of captions into Hindi. Due to time and financial constraints, machine translation is being used in this project. The greatest machine translation software is Google Cloud Translator, which offers a free API to translate approximately 71 languages.

Google's English-Hindi translation received an Assessment of Text Essential Characteristic (ATEC) of 0.402. It is one of the highest ratings ever obtained in an English to Hindi computer translation (Malik and Baghel, 2019). Thus, 40,000 sentences were translated into Hindi language by utilizing the Google cloud translation API. The caption dataset sample for a picture in English and Hindi is shown in Figure 7 below.

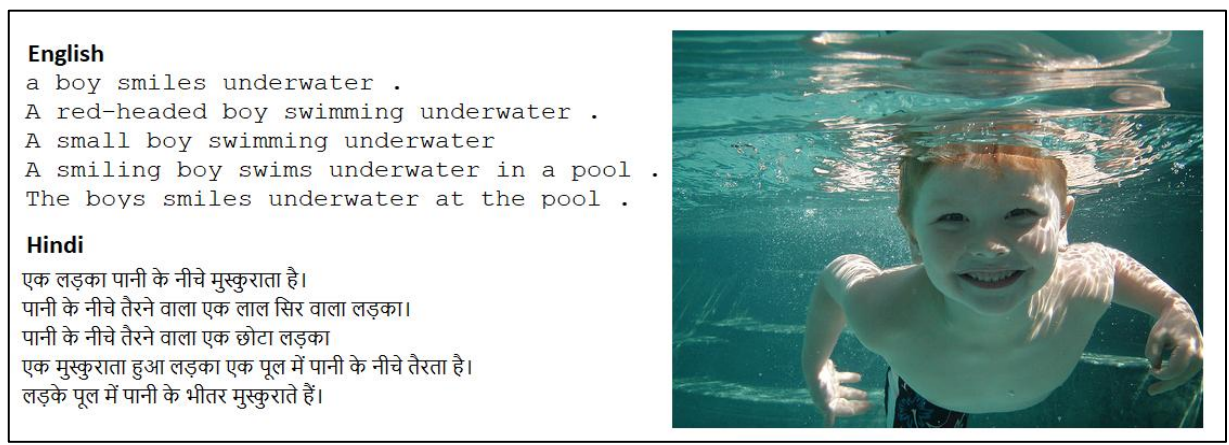

*Figure 7: Example of English and Hindi captions*

### 4.4 Data Processing

The 8,000 images in the Flickr8k-Hindi collection are made up of both text and image data (40,000 sentences). Before being fitted into the model, both data need to be pre-processed and transformed. The preparation of textual data and images, before transmission to neural network picture caption-generating model, is covered in this section.

### 4.4.1 Text data

The sole component that needs to be predicted in an image captioning model is the text. Text data will therefore be goal variable which machine understands to predict during training. The textual data must therefore be processed and turned into a numeric data before being fed into the neural network. More information on the pre-processing and text data transformation utilised in the model for image captioning is provided in this section.

The five captions for each image are provided from the flickr8k-hindi dataset, as was mentioned in the section above. The Flickr8k-Hindi Dataset has a total of 10,889 distinct words, with the Hindi word "ek," which meaning "a" or "one," being the most often used word. Because it contains little information, this word can be eliminated, which will help the model perform better by reducing the vocabulary. The fluency of the produced caption may suffer if stop words are removed (Lan et al., 2017). As a result, just stop words remain in the text and all punctuation and numerical texts have been removed. Once cleaning up the description data is accomplished, the vocabulary's text count was reduced to 10,870, which is still a significant amount. One caption is therefore removed from the total of five captions in order to reduce the vocabulary size.

Following pre-processing, text data is marked with start and end markers to instruct the computer that these are the initial part of a sentence and final part of a sentence, respectively. The illustration displayed in Figure 8 below. As well, Project Design flow already describes the significance of these tags.

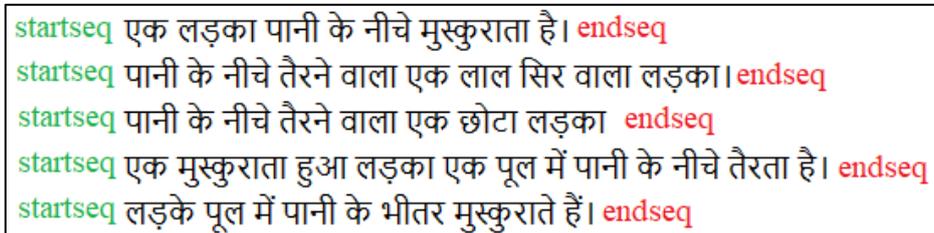

*Figure 8: Hindi caption wrapped with marker tags*

The text cannot be directly processed by the neural network. Therefore, using the Keras tokenizer, each word is changed into an integer token once the phrase has been labelled as previously indicated. However, as the neural network cannot process integers either, the vectors of floating point values is created using Keras embedding layer. In order for them to be combined and managed as fed to the deep learning decoder model, vector embedding size is set as identical to image vector feature by doing padding to zero.

### 4.4.2 Image data

When feeding an image into a deep neural network, it must take the shape of a vector. Hossain et al. (2018) conducted a survey on "Deep learning for image captioning" and found that every state-of-the-art study on image captioning employed pre-trained CNN to turn images into fixed-sized vectors. For the purpose of creating picture captions, there are numerous pre-trained CNN models available, including GoogleNet, AlexNet, ResNet-50, Inception-V3, and VGG-16 (Bai and An, 2018). VGG-16, Inception-V3, and ResNet-50 have demonstrated the significant outcome for the flickr8k dataset amongst these pre-trained Convolutional neural networks (Bai and An, 2018). Additionally, CNNs which are pre trained prevents overfitting in the model for captioning the image (Vinyals et al., 2015).

Final layer of the pre-trained CNNs are typically used to forecast image's classification. In order to build the model, the final layers of the pre trained CNNs has been deleted. This is being replaced with the penultimate layer, that sends picture features as vectors. ResNet-50, VGG-16, and InceptionV3 photo features are saved into the pickle rather than processing image features during model run each time. This will speed up and use less memory during training of the model for image captioning. As a result, three image feature files are produced for the Inception-V3, VGG-16, and ResNet-50 algorithms.

## 4.5 Model Development

In this study, numerous models with various pre-trained CNNs were employed to run numerous tests. We will talk about the training model in this part that produced the best captions for the test image. Research utilizes easy to use pre-trained CNN which generate good output and consume comparatively smaller amount of training time. In contrast to self built Convolutional neural networks, pre trained CNNs generate more precise outputs.

When the images features are extracted, there could be an issue of containing the scenes of the image in the caption. The LSTM is employed to resolve this issue. Accuracy of produced text precision is restricted to brief words. In order to resolve this restriction Bi-LSTM is employed. Bi- LSTM is excellent while lengthier sentences are utilized as it maintains additional data.

### 4.5.1 Text Elements

The initial file to be fed into the procedure is a description file of images. An ID of an image with a listing of five captions which explain the picture is included in the description file. This file was then processed and saved into a dictionary, with the image ID serving as the key and the image description serving as the values. A text file containing these pre-processed descriptions is later created. The text encoder uses these inputs as a second input. The highest length of description is 39 characters; hence 39 characters should make up input sequence. 0.5 is defined as a dropout rate in the dropout layer, is next layer to process the input sequence after the encoding layer. Both uni-LSTM and bi-LSTM layers are used to train and test the mode.

### 4.5.2 Image Elements

VGG16 delivered the best outcomes out of all the pre-trained CNN models that were used. The Keras library was used to import the pre-trained CNN. The CNN was afterwards launched and reorganised. The CNN was restructured by deleting its last layer. Since the last layer is a classification layer, it was taken out of the model. The desired size for the photos is set to 224*224. Before they are moulded for the model, the image pixels are recorded in a numpy array. These procedures are followed by the extraction of picture features. In order to use the extracted picture features later without having to repeat the feature extraction procedure, they are saved in a pickle file. The input structure of encoder of the image and the feature of the image are same because VGG16 produces an image vector of 4096 size of array built on the final layer, this is supplied as to image encoder.

### 4.5.3 Combination of Text and Image Elements

Both the text and image encoders provide input to the decoder. The two encoders are combined using the add operation. The dense layer, which uses 256 neurons for both bidirectional LSTM and unidirectional and receives this next. SoftMax is utilized in the dense layer as an activator. Adam is used in compiling the model when each step has been finished.

### 4.5.4 Attention Layer

In image captioning same caption may repeat for several pictures. An example could be, in case an image is about a boy performing any action instead of a walk, in that case a model produces a description as "Boy is walking on the road". Such captions are repetitive for various pictures that are linked to a boy. The explanation behind this is that the models employ approach such as greedy search that expects the next possible phrase to be conditioned on the feature of an image and earlier word.

There is maximum chance of overfitting of the data and the key words that occur in a smaller amount in text and would have lower likelihood. In order to prevent this constraint, this research employs Attention mechanism which utilizes similar encoder and decoder-based modelling which rather generates additional weights to the appropriate text and image. Attention layer helps to generate a weight vector and, by multiplying it, combine image characteristics from each time step into a sentence-level feature vector.

### 4.5.5 Data Loading

Our image collection requires a lot of RAM to process because of its size. The system used to run this model doesn't have enough memory to process all of the data at once. The data is incorporated into the model gradually as a solution to this problem. A generator function that creates single dataset batch is written. These dataset batches are sent to the model during fitting. Keras facilitates data loading in stages.

### 4.5.6 Model Tuning

With the knowledge gained from the experiment, this project found the ideal hyperparameter faster by using a smaller sample of the larger dataset. Number of images are two hundred with five descriptions each make up the dataset's small sample. The test dataset and training dataset both receive an equal portion. As the neural network model nature is more contingent, alteration in the model is investigated, by running the model numerous times, an ideal number of models run repeat is discovered. The test harness was consistent after five model run repetitions, according to the model evaluation. As a result, a thorough series of experiments is carried out to evaluate the model's effectiveness by using various model architectures in order to determine the ideal hyperparameter.

## 4.6 Experiments

In this study various experiments with different models are employed but the structure of the best model is highlighted which produces the better quality of image captions in Hindi. The experiments are done with LSTM, Bi-LSTM and Att-BiLSTM to have a comparison with the with different approaches. After these experiments, a comparative analysis of the results will be done to highlight the importance of the study.

In each experiment models are trained with different pre-trained CNN models with Unidirectional LSTM, Bidirectional LSTM, and Bidirectional LSTM with Attention mechanism which provides extra weights to the appropriate text and image. To avoid overfitting dropout rate is also considered. The value of this rate is finally considered as 0.5. A loss function of Softmax cross-entropy is used to make the model output be as close as possible to the desired output as in model training, the model weights are iteratively adjusted accordingly with the aim of minimizing the Cross-Entropy loss. Furthermore, the results of the Adam optimizer are generally better than any other optimization algorithms, and has faster computation time, and require fewer parameters for tuning. Because of all these reasons, Adam is recommended as the default optimizer and used in model training.

### 4.6.1 Experiment 1: Using VGG16

The pre-trained Convolutional neural networks used in the first experiment for feature extraction was VGG16. To extract features from the photos, Att-BiLSTM, Bi-LSTM, and LSTM were all used. These models were created to function in a variety of epochs.

The top BLEU1 score for LSTM was 0.55, followed by the best BLEU1 scores for Bi-LSTM and Att-BiLSTM, respectively. Although there was no discernible difference in the BLEU score, it was clear from the captions that the Att-BiLSTM model produced a more accurate description of the image. The figure 9 below shows that all LSTM models were successful in correctly identifying the objects in the photos. In the image, there are no undesired objects that the Att-BiLSTM model could detect. The caption for the Att-BiLSTM model makes more sense when the image's context is taken into account.

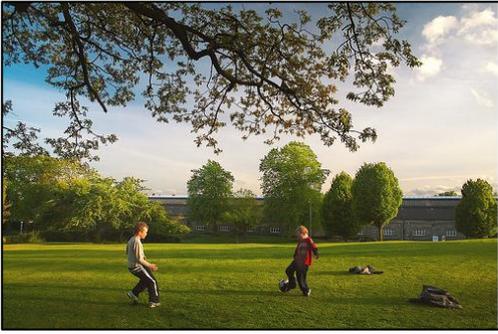

*Figure 9: Caption generated by VGG16*

### 4.6.2 Experiment 2: Using InceptionV3

The features of the image were extracted in the second experiment using InceptionV3. Like previous experiments, this experiment to use three LSTMs as an encoder. The LSTM model yielded a very low BLEU1 score of 0.38, while there was significant update to Bi-LSTM model BLEU1 score of 0.48. No significant improvement was made as Att-BiLSTM model yielded a BLEU1 score of 0.49. The experiment produced relatively subpar results because the model is unable to extract details or features from the image.

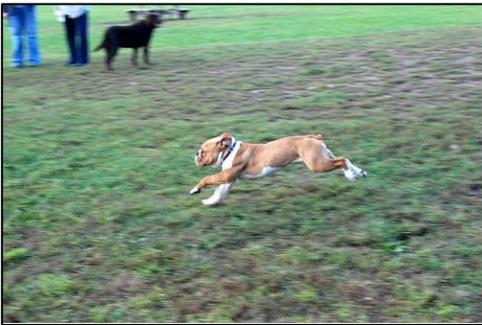

*Figure 10: Caption generated by InceptionV3*

### 4.6.3 Experiment 3: Using ResNet50

In the third experiment, image features were extracted using a ResNet50 pre-trained CNN. For this model, text encoding was done using LSTM, Bi-LSTM, and Att-BiLSTM. The experiments were planned to last for a variety of epoch counts. The BLEU1 score for the LSTM model was 0.52, the BLEU1 score for the Bi-LSTM model was 0.53, and the BLEU1 score for the Att-BiLSTM model was 0.56. In both models, BLEU score differences were not statistically significant, unlike the VGG16 model. Even with this model, as seen in Figure 11 below, both models produce a caption that describes the man and the rocks in the background of the picture.

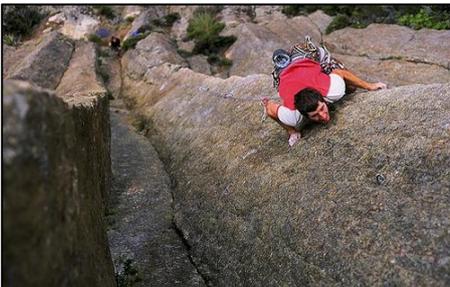

*Figure 11: Caption generated by ResNet50*

However, the Att-BiLSTM model also predicts an action, like climbing and ropes in the background mentioned in the captions, which gives the image context. All LSTM models anticipate an undesired object that is not in the image.

## 4.7 Summary

This section focused on the technical steps that were taken to put the study of Hindi image captioning into practice. The details of setting up the environment for training the models were highlighted precisely in this section. Specification with respect to data, data preparation, pre-processing and transformation were also underlined. Different approaches were talked about and how these approaches were implemented in various experiments.

This section emphasizes what all special touches and features were implemented for the study. The focus is more different settings that were carried out during the training of the models. The details of model implementation, structure of layers parameter tuning, employment of pre-trained CNNs were also highlighted and brought forward in this section. Sample of output were also shared in this section, but more details and discussions of the results will be carried out in the next section.

## 5. RESULTS AND DISCUSSIONS

This Section includes the result of the models trained on Flickr8k Hindi datasets. The trials that were carried out for tweaking yielded the model's ideal parameter. After explaining the key project findings, the detailed result analysis is carried out and comparison with previous works is discussed.

While the quality of the captions which are created, are able to be assessed manually, a subjective rating is even necessary to indicate their level of quality. At this point, the caption is assessed while taking into account the reference descriptions. The research employs BLEU scores, which describes the worth of the generated caption (Papineni et al., 2002). BLEU score is a widely used evaluation method in image captioning. There are 8000 pictures in the test set; at this point, every image corresponds to five captions which define object actions in the picture. In this section calculation models descriptions of BLEU-1, BLEU-2, BLEU-3, and BLEU-4 scores are compared to another baseline.

In this section, the detailed discussion of using different pre-trained CNNs and their results are highlighted. This helps to understand which CNN model will yield the best results and can help us to get better caption s in Hindi for the images. A high-quality caption comparison is also done further with the images to give the study more highlights of the approaches that were applied. One of the important aspects of the result is signified by the use of Attention based mechanism which helps to compare the results with other BI-LSTM models.Moreover, the in-depth details of parameter tuning are discussed in this section which helped to achieve the better BLEU scores. Some previous works on Hindi image captioning are also discussed and precise comparison is also done with this research.

## 5.1 Bilingual Evaluation Understudy Score of Experiments

The original purpose of the BLEU (Bilingual Evaluation Understudy) was to assess machine translated sentences. It is employed to gauge how similar the candidate sentence and the reference sentence are to one another. By tracking the co-occurrences of n-grams, it evaluates the generated description's quality in relation to multiple reference descriptions. The fluency of the resulting description is evaluated using the lengthier n-gram scores (Papineni et al., 2002). Since using one to four n-grams is the most typical approach (Bai and An, 2018), this research has used one to four n-grams. BLEU-4 gauges the description's fluency, while BLEU-1 (unigram) assesses its sufficiency. Additionally, it assesses the resulting description's precision, which is calculated as the number of overlapping words to overall number of words in the contender phrase (Papineni et al., 2002). The best way for assessing the characteristic of the produced image captions is by human evaluation (Elliott et al., 2016). Time constraints prevent manual picture description evaluation, and human evaluation is not taken into account.

Different experiments are carried out for the study and implementation was discussed in detail in the previous section. It is evident from the sample captions from the experiments that Attention mechanism based Bi-LSTM models performed better and provided improvements to Bi-LSTM models. The attention mechanism is introduced to improve the performance of the encoder-decoder model for machine translation. Also, pre-trained CNNs like VGG16, InceptionV3 and ResNet50 had different scores. These CNNs produced scores based on different technical setups that are implemented for model training

The BLEU score for experiments is displayed in Table 1 below. The InceptionV3 LSTM model got the lowest BLEU1 grade of 0.38 whereas the VGG16 Att-BiLSTM model had the best grade of 0.59.

*Table 1: BLEU Score of Experiments*

| Dataset | Pre trained CNNs | Baseline | Best Epochs | BLEU | | | |
|---|---|---|---|---|---|---|---|
| | | | | BLEU 1 | BLEU 2 | BLEU 3 | BLEU 4 |
| Flickr8k-Hindi Dataset | VGG16 | LSTM | 20 | 0.55 | 0.37 | 0.25 | 0.12 |
| | | CNN-BiLSTM | 10 | 0.56 | 0.42 | 0.29 | 0.12 |
| | | CNN Att-BiLSTM | 20 | 0.59 | 0.43 | 0.29 | 0.19 |
| | InceptionV3 | LSTM | 10 | 0.38 | 0.23 | 0.13 | 0.10 |
| | | CNN-BiLSTM | 10 | 0.48 | 0.37 | 0.21 | 0.11 |
| | | CNN Att-BiLSTM | 10 | 0.49 | 0.39 | 0.22 | 0.12 |
| | ResNet 50 | LSTM | 20 | 0.52 | 0.41 | 0.28 | 0.11 |
| | | CNN-BiLSTM | 10 | 0.53 | 0.37 | 0.25 | 0.12 |
| | | CNN Att-BiLSTM | 10 | 0.56 | 0.42 | 0.26 | 0.14 |

**5.1.1  Experiment 1: VGG16 BLEU scores:**

In this experiment pre-trained CNN VGG16 was employed for LSTM models of Uni-LSTM, Bi-LSTM and Att-BiLSTM. For VGG16 LSTM, BLEU-1 score was 0.55 which is still very considerable as it can provide meaningful caption to the image. Using Bidirectional LSTM with VGG16 gets insignificant BLEU-1 score of 0.56 but adding an Attention layer to the Bi-LSTM improves the score significantly to 0.59 at 20 epochs being considered for getting the good quality score. This demonstrates that compared to previous LSTM models and other tests, the Att-BiLSTM model has produced more contextually meaningful and informative phrases. The investigation revealed that the VGG16 Att-BiLSTM model was the best one. No undesirable things in the image were picked up by the model. The caption for the Att-BiLSTM model makes sense when the context of image is studied.

**5.1.2  Experiment 2: InceptionV3 BLEU scores:**

InceptionV3 is prominently used for assisting in image analysis and object detection but could not provide more convincing results in this research. When used with LSTM models low BLEU1 score were recorded for InceptionV3. For InceptionV3 LSTM, BLEU-1 score was 0.38 which is the lowest in all the experiment. Using Bidirectional LSTM with InceptionV3 improved the score significantly to 0.48 but adding an Attention layer to the Bi-LSTM had little impact by logging BLEU1 score of 0.49. Moreover, 10 epochs were considered in every approach o this experiment as increasing the epochs degraded the scores. It is very evident that image captions generated with this experiment are occasionally inappropriate.

### 5.1.3 Experiment 3: ResNet50 BLEU scores:

For this experiment, ResNet50 Convolutional neural network was utilized. Again, ResNet50 was employed with LSTM models for feature extractions. The image caption generated for this experiment were contextually correct. Further down the experiment, scores soared, and also adequacy and fluency of captions were reasonable. BLEU1 score for ResNet50 LSTM was 0.52 over 20 epochs. As the epoch was reduced to 10, the score increased considerably for Bi-LSTM to 0.53. There was significant upgrade to score when for Att-BiLSTM and caption generated were understandable.

### 5.2 Interpretations of Findings

The outcomes of the approaches accomplished in the course of model tuning is provided in the prior segment. The best captions are the one which generates accurate and descriptive explanations what the image is trying to depict. Figure 12 below shows generated captions.

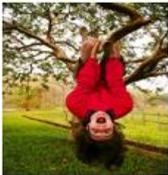
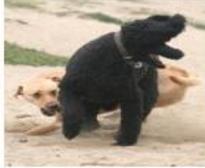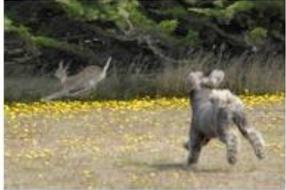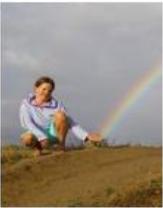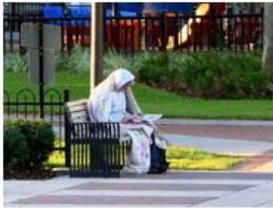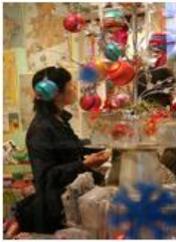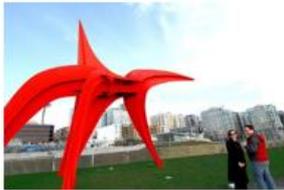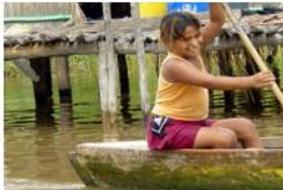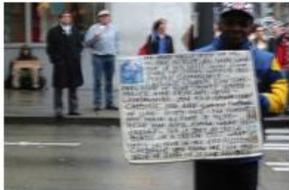

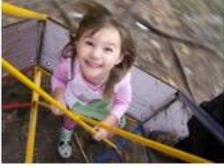

*Figure 12: Image captions results*

### 5.2.1 Result Analysis

From the results depicted by BLEU scores, it is evident that when a BiLSTM-RNN are used with pre-trained CNNs, there is minor improvement in the scores. But it does not assure that Bi-LSTM models will always produce a higher score. A very remarkable thing to note is that when an attention layer is added to the Bi-LSTM it enhances the scores significantly and produce a superior result. Following are the key discoveries of this study:

- Models trained with Att-BiLSTM have a higher BLEU score than those trained with LSTM and Bi-LSTM models (Table 1). The next likely phrase is predicted using a greedy search by unidirectional and bidirectional LSTM models based on picture feature and prior research. The terms with lower likelihood are those that appear less frequently in the text, and this technique soon overfits the data. Using the same encoder-decoder model, the attention mechanism gives the pertinent image and text more weight.

- A model trained with the pre-trained CNNS like VGG16 and ResNet50 produces a very good quality captions compared to InceptionV3 model trained different approaches.
  Figure 12 and 13 below shows the Good, Average and Poor-quality image captions generated by well performing models.

- Out of five models above, VGG16 Att-BiLSTM model generates a very good quality image caption. Moreover, it also expends comparatively a smaller amount of memory when compared to other models.

### 5.2.2 Error Analysis

When the generated captions were incorrect, they frequently displayed flaws that broadly belonged to the following groups: error in classification, error in numbering, error in colour identification, error in gender recognition, error in presence of object. Below are the definition and summary of these mistakes.

- Error in classification: This error exists when model incorrectly recognises the object, animal, or human in the picture.

- Error in numbering: This error exists when model is not able to determine the count of the object, animal, or human in the picture.

- Error in colour identification: This error exists when model is not able to determine the accurate colour of the object or animal in the picture.

- Error in gender recognition: This error exists when model is not able to recognise the gender recognition of a person in the picture.

- Error in occurrence of object: This error exists when model is not able to determine whether the object, animal, or human occur in the picture.

The prior examples demonstrate that some of these mistakes can be trivial and only result in captions that are only partially accurate. But when the models try to caption the pictures, certain mistakes can result in inconsistencies or absurd claims.

## 5.3 Comparison with Previous Works

The analysis of created captions and their qualitative and quantitative findings is covered in this part. As far as it is known, there is very limited research on Hindi image captioning (Dhir et al., 2019). As a result, the strategy is to do a comparative study with this available research. Additionally, this research contrasted our findings with a few well-known encoder-decoder designs that are routinely applied to image captioning. The researchers used the MS-COCO dataset, and Google Translate was used to translate the description.

*Table 2: Dhir et.al. BLEU score*

| State of Arts | BLEU Scores | | | |
|---|---|---|---|---|
| | B1 | B2 | B3 | B4 |
| Resnet 101 and GRU | 57.0 | 39.1 | 26.4 | 17.3 |
| Baseline 1 | 56.4 | 38.8 | 26.3 | 17.2 |
| Baseline 2 | 56.3 | 38.4 | 25.8 | 16.9 |
| Baseline 3 | 55.7 | 38.4 | 25.9 | 16.8 |

To preserve the descriptions quality, the researchers engaged two human translators to edit it manually and review the captions. Their BLEU1 score came out to be 0.57. Transformer networks were employed by Mishra et al. (2021) to produce Hindi captions. Using Google Translate, the experiment's dataset was produced. The MS COCO dataset was utilised by the authors for translation. The limitation in our research is human evaluation was not considered. Despite of this our model yielded good score. Dataset in this study also differs to the dataset we are using. Hence the data preparation and translation may also differ. Other study by (Mishra et al., 2021) involved MS-COCO dataset. This study tried to fill some gaps in previous models on image captioning. This study employs encoder-decoder based transformer architecture for image captioning. Table 3. Below shows the BLEU scores of the research.

*Table 3: Mishra et al. BLEU score*

| Author | BLEU Scores | | | |
|---|---|---|---|---|
| | B1 | B2 | B3 | B4 |
| Mishra et al. | 62.9 | 43.3 | 29.1 | 19.0 |

## 5.4 Discussion

The results of study were compared by using BLEU scores. For the purposes of comparative analysis of this research, many experiments were conducted which act as a baseline. A score of BLEU-1 indicates that the image is adequate, and a score of BLEU-4 indicates that the image captioning is fluent. The attention-based bidirectional LSTM with VGG16 generated the greatest results for both BLEU scores. The best scores in model training across all experiments were 0.59 and 0.19 respectively.

Bidirectional LSTM models generated superior quality captions and offered a greater BLEU score as compared to other LSTM models. The Bi-LSTM models produced contextually significant and instructive phrases than the unidirectional LSTM models by adhering to its specification. It is evident that approach of greedy-search which forecast the following likely phrase prepared on image feature and prior phrases was handled when the Attention layer was apprehended to the Bi-LSTM models. This helped to reduce overfitting of data and generate good quality captions.

Six models were compared, and the model that was trained with a VGG16 Att-BiLSTM generated high quality captions. However, for some models, BLEU score was poor, this could be because there are not as many reference words to match. Since lesser research are conducted to produce Hindi image captions, and the outcome of the research is crucial in image captioning field. Results of this research may be helpful for future investigations on Hindi image captioning.

According to(Vinyals et al., 2015), evaluation done by humans can be considered as the best way to assess the image captions, and the findings of this study seem to support this claim. Several examples from the experiments are classified as good, average, and poor. This helps to understand the performance of the model by comparisons. The Att-BiLSTM model has produced more contextually significant and instructive sentences than the other LSTM models by adhering to its definition.

Additionally, model trained with the VGG16 Att-BiLSTM had the BLEU score which is comparable to that of the model trained with the Hindi captions by (Dhir et al., 2019) and(Mishra et al., 2021). It has also been demonstrated in other studies that crowdsourced descriptions provide captions that are more naturally occurring. However, because of time constraints, study is unable to gather the image description from the Hindi native speakers. The output of the experiments may prove helpful for subsequent work that makes use of crowdsourced captions.

## 5.5 Summary

This section was dedicated to discussing the insight of results extracted out of the experiments conducted. In this section results were evaluated, and in-depth analysis of the results was done. Results from various experiments were thoroughly compared to verify the picture captions quality. How model tuning and experimental settings helped to gain better results were also reviewed in this section. The motive of this section is to explain and justify what results indicates and how can this result help us to decide in the research.

This section also emphasizes how this research has accomplished as compared to other previous studies. The brief discussion is carried out to measure the performance based on result analysis. This section tries to convey the analysis of the outcomes carried out to generate image captioning in Hindi.

# 6. CONCLUSIONS AND RECOMMENDATIONS

This section provides the discussion and summary of the research of Using deep learning to generate semantically correct Hindi captions. This section consist of synopsis of how this research was carried out and how it differs from previous works. The research gaps in previous works were identified in the field of image captioning and were implemented. This implementation is also talked about in this section.

In this section, contribution to knowledge is also discussed as how research was conducted and what ways, approaches and techniques were utilized. It highlights the major achievements of image captioning in Hindi Language. How this research tried to achieve something different. In this section it is underline that how this research can contribute to further studies and works.

Future recommendations to the field of deep learning and image captioning are shared in this section. Discussion about previous contrast in studies and how this research can be utilized for image captioning in other languages are mentioned further. This research can act as a base to other works and any amendments are welcomed. Future research in the field of Neural networks and image captioning in Non-English languages can be benefited from this research.

## 6.1 Conclusion

This study offers well-generated Hindi captions for the images, which will be helpful to a significant portion of the population. At the time of this study, there are about 500 million hindi speaking people in the world. Generating good captions can aid different section of this Hindi speaking people. As visuals have become a vital element of our daily lives because of evolving technology. Every single day, millions of images are shared online. The importance of image captioning rests in the way it increases access to the media. This study will support initiatives of other international studies on picture captioning. Another area where this research can be useful is in field of education. Children who are native Hindi speakers can be introduced to a variety of ideas that may be useful for early learning. It would be beneficial to keep an eye on or carefully review interactions on various social media platforms to abide by the regulations. With the help of skilled captioning models, the crisis management teams can ascertain the underlying circumstances to react appropriately.

This study shows how encoder-decoder models can help in Hindi image caption creation. We can distinguish between the level model performance using both LSTM and Bi-LSTM. When Attention layer was added to Bi-LSTM, Att-BiLSTM models created better image captions and had a superior BLEU score when compared to the other LSTM models. The Att-BiLSTM model produced additional contextually significant and instructive captions than other LSTM models by adhering to its definition. This research signifies the importance of Attention mechanism which helps combine image characteristics and extracts more insights from the image.

BLEU scores were used compare the outcomes of the study. Different experiments were performed as a baseline for this research to have a comparative analysis. Performances of the models were noted across different experiments. BLEU-1 score signifies the adequacy of the image and BLEU-4 denotes the Fluency of image captioning. For both the BLEU scores Attention based Bidirectional LSTM with VGG16 produced the best results. Score of 0.59 and 0.19 were recorded as the best score in model training for all experiments.

In this study, the sample showed how the image caption generated for the best five models. The evaluation done by BLEU could have been more assisted by human evaluation which could not be done because of time constraint. This research encourages to be used as a base to further studies and remains firm to focus on generating the best defining captions in Hindi.

## 6.2 Contributions

By utilising machine translation to collect 40,000 captions for 8,000 photos, this study has created a dataset for Hindi image captions. With the use of these datasets, an encoder decoder deep learning model can be trained which instinctively see a picture and produce a comprehensible Hindi caption. After acquiring image feature and previous word, the training of the model is carried to predict words.

Various Bi-LSTM model experiments revealed that using this dataset to train the model improved the quality of image captioning. The studies show that the model produces very good quality captions when trained with a single cleaned description for each image. Additionally, a model trained using CNN VGG16 Att-BiLSTM achieved the greatest BLEU score, which is the cutting-edge outcome for describing an image in Hindi. The findings of this study can be utilised as a starting point for future research and to add to the body of knowledge on Hindi image captioning.

## 6.3 Future Recommendations

In this research we train the image captioning model with sentences that have been automatically translated. The translated sentences do not necessarily need to be accurate to the nth degree by the machine translator. Due to the differing grammatical structures of the languages, it is difficult to translate sentences from one language to another. Experiments are conducted taking this machine translator constraint into consideration. It has also been demonstrated in other studies that crowdsourced descriptions provide captions that are more naturally occurring. However, because of time constraints, study was unable to gather the captions from the Hindi native population. The output of the experiments may prove helpful for subsequent work that makes use of crowdsourced captions.

Also, human evaluation can help to refine and evaluate the results. The generated captions with high BLEU scores should be gathered for human evaluation, and every single produced caption should be evaluated by looking at the pictures. This can assist to enhance the model performances in future. Human evaluation will also give an edge to distinguish the ability model to produce better caption generation.

Different data set such as MS-COCO or combination of MS-COCO, Flickr8K and other available datasets can be utilized for the future studies. Crowd sourcing image captioning can also be looked at as it proves good source of data directly from the speaker. GRU models instead of LSTMs can be looked upon to create baselines for potential studies. In addition, other pre-trained models could be leveraged for better feature extractions. The development of a dense model for image captioning in Hindi, where several captions are created for each image. This work can be expanded upon for the creation of paragraphs from images.